\documentclass[letterpaper, 10 pt, conference]{ieeeconf}  

\IEEEoverridecommandlockouts                              

\usepackage{graphics} 
\usepackage{epsfig} 
\usepackage{subfigure}
\usepackage{amsmath} 
\usepackage{epstopdf}
\usepackage{hyperref}
\usepackage{lipsum,amsmath,multicol}
\usepackage{float}
\usepackage{algorithm}
\usepackage{algorithmicx}
\usepackage{algpseudocode}

\title{\LARGE \bf
Trajectory Optimization for Curvature Bounded Non-Holonomic Vehicles: Application to Autonomous Driving
}

\author{Mithun Babu$^{1}$, Yash Oza$^{1}$, C. A. Balaji$^{1}$ Arun Kumar Singh$^{2}$, Bharath Gopakarishnan$^{1}$, K. Madhava Kirshna $^{1}$
\thanks{The research was partly supported by TUT postdoc fellowship to the first author}
\thanks{
$^{1}$ Robotics, Research center, IIIT Hyderabad
$^{2}$ Department of Hydraulics and Automation, Tampere University of Technology, Finland
      }
}

\begin{document}

\maketitle
\thispagestyle{empty}
\pagestyle{empty}

\begin{abstract}
In this paper, we propose a trajectory optimization for computing smooth collision free trajectories for non-holonomic curvature bounded vehicles among static and dynamic obstacles. One of the key novelties of our formulation is a hierarchal optimization routine which alternately operates  in the space of angular accelerations and  linear velocities. That is, the optimization has a two layer structure wherein angular accelerations are optimized keeping the linear velocities fixed and vice versa. If the vehicle/obstacles are modeled as circles  than the velocity optimization layer can be shown to have the computationally efficient \emph{difference of convex} structure commonly observed for linear systems. We build upon this insight to extend the trajectory optimization to polygonal obstacles. In particular, we use the Minkowski sum and the circumscribing circle of the resulting polygon to reduce collision avoidance between a pair of  convex polygons to that between a point and a circle. This leads to a less conservative approximation as compared to that obtained by approximating each polygon individually through its circumscribing circle. We use the proposed trajectory optimization as the basis for constructing a Model Predictive Control framework for navigating an autonomous car in complex urban scenarios like overtaking, lane changing and merging. Moreover, we also highlight the advantages provided by the alternating optimization routine. Specifically, we show it produces trajectories which have comparable arc lengths and smoothness as compared to those produced with joint simultaneous  optimization in the space  of angular accelerations and linear velocities. However, importantly,  the alternating optimization provides some gain in computational time.
\end{abstract}

\section{Introduction}
Trajectory optimization of non-holonomic vehicles is challenging owing to the highly non-linear motion model. Moreover, incorporation of hard bounds on curvature further increases the complexity of the optimization. In fact, we remark that it is the presence of hard curvature bounds which acts as a key bottleneck of the problem. To understand this further, consider the following two equivalent discrete time representations of a non-holonomic vehicle with curvature bound constraints.

\begin{figure}[H]
  \centering
   \subfigure[]{
    \includegraphics[width= 4.25cm, height=3.0cm] {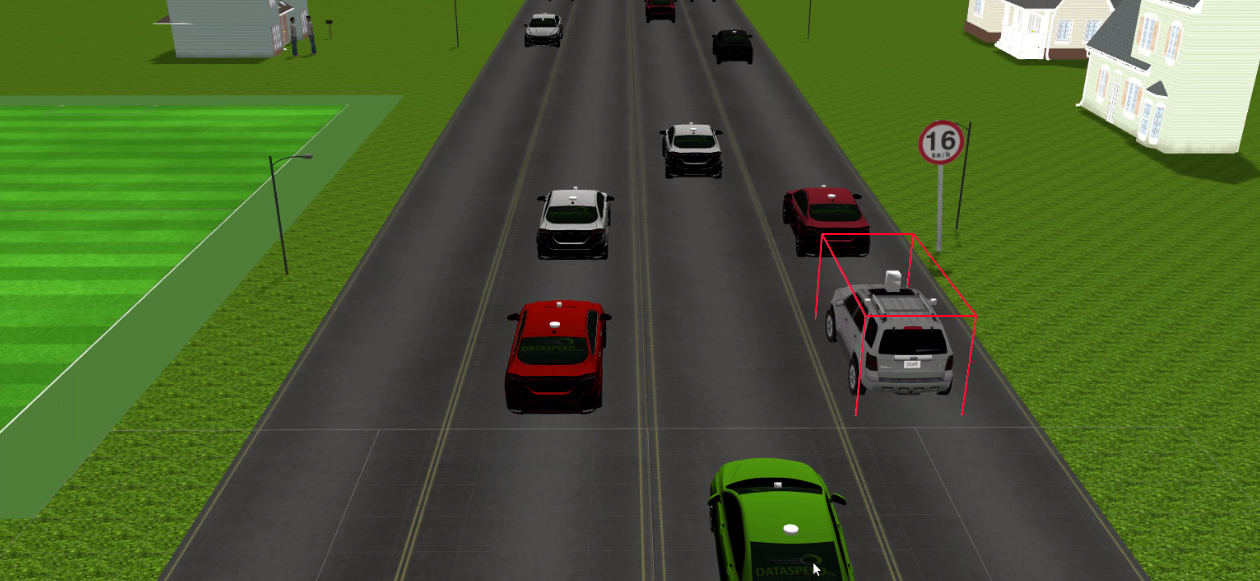}
    \label{ex_plot1}
   }\hspace{-0.6cm}
   \subfigure[]{
    \includegraphics[width= 4.25cm, height=3.0cm] {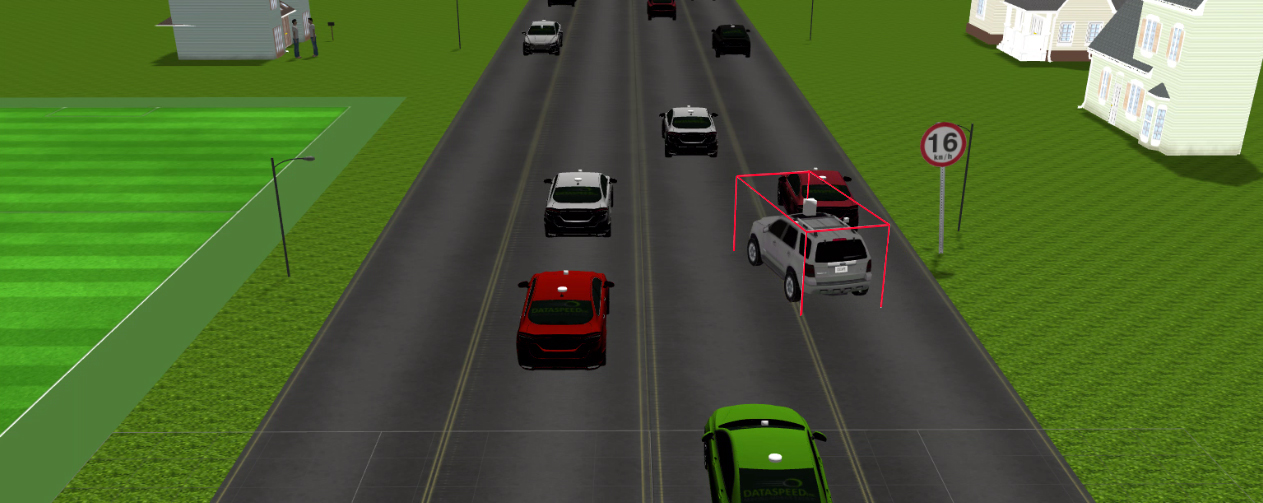}
    \label{ex_plot2}
   }
   \subfigure[]{
    \includegraphics[width= 4.25cm, height=3.0cm] {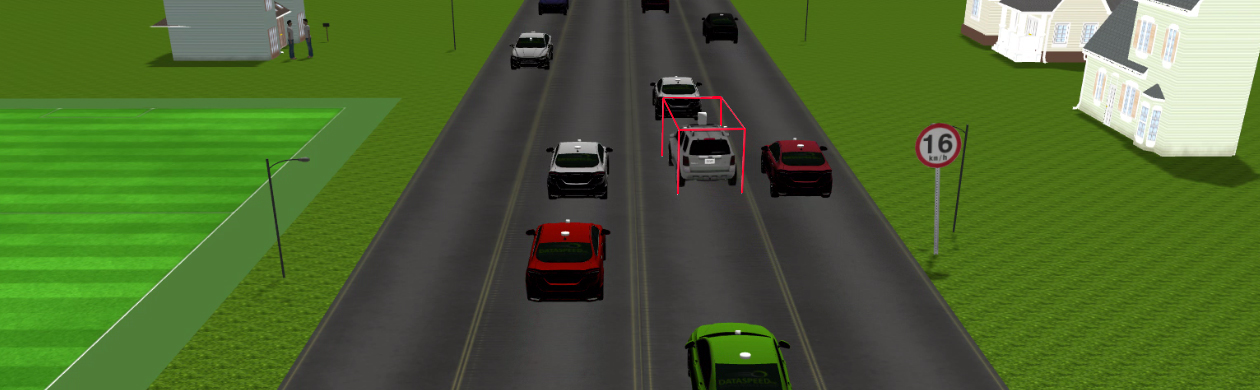}
    \label{ex_plot2}
   }\hspace{-0.6cm}
   \subfigure[]{
    \includegraphics[width= 4.25cm, height=3.0cm] {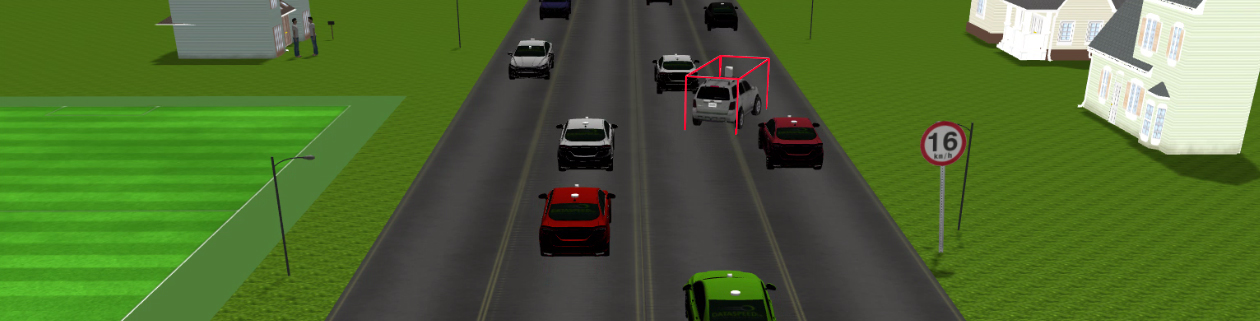}
    \label{ex_plot2}
   }
   \caption{Autonomous vehicle (shown in red box) performing overtaking maneuvers based on a MPC framework built using the trajectory optimization proposed in the paper.}          
\end{figure}

\begin{equation}
  \begin{cases}
    x(t_{i+1}) = v(t_i)\cos\theta(t_i)\Delta t.\\
    y(t_{i+1}) = v(t_i)\sin\theta(t_i)\Delta t.\\
    \theta(t_{i+1}) = \theta(t_i)+\dot{\theta}(t_i)\Delta t+ \frac{1}{2} \ddot{\theta}(t_i)\Delta t^2\\
    \dot{\theta}(t_{i+1}) = \dot{\theta}(t_i)+\ddot{\theta}(t_i)\Delta t\\
    \ddot{\theta}(t_i) = u_1(t_i)\\
    v(t_i) = u_2(t_i).\\
    -\kappa_{max}v(t_i)\leq \dot{\theta}(t_i)\leq \kappa_{max}v(t_i).  
  \end{cases}
  \label{nonhol1}
\end{equation}

\begin{equation}
  \begin{cases}
    x(t_{i+1}) = x(t_i)+\dot{x}(t_i)\Delta t+ \frac{1}{2} \ddot{x}(t_i)\Delta t^2.\\
    y(t_{i+1}) = y(t_i)+\dot{y}(t_i)\Delta t+ \frac{1}{2} \ddot{y}(t_i)\Delta t^2.\\
    \theta(t_{i}) = \arctan2(\frac{\dot{y}(t_i)}{\dot{x}(t_i)})\\
    v(t_i) = \sqrt{\dot{x}(t_i)^2+ \dot{y}(t_i)^2}.\\
    \ddot{x}(t_i) = u_1(t_i).\\
    \ddot{y}(t_i) = u_2(t_i).\\
    -\kappa_{max} \leq \frac{\ddot{y}(t_i)\dot{x}(t_i)-\ddot{x}(t_i)\dot{y}(t_i)}{(\dot{x}(t_i)^2+\dot{y}(t_i)^2)^\frac{3}{2}}\leq \kappa_{max}
  \end{cases}
  \label{nonhol2}
\end{equation}

\noindent Where, $x(t_i), y(t_i), \theta(t_i) $ are the position and heading of the vehicle at time, $t_i$ while $u_1(t_i), u_2(t_i) $ are the control inputs at some instant $t_i$. The terms, $\kappa_{max}, \kappa_{min}$ represent the maximum and minimum curvature bounds respectively. Thus, the inequalities in both the representations define the curvature bound constraints. The representation (\ref{nonhol1}) is the classic non-holonomic model while (\ref{nonhol2}) is obtained via feedback linearization \cite{rufli}.  In (\ref{nonhol1}), the curvature constraints are affine functions of control inputs. Thus, a trajectory optimization which uses (\ref{nonhol1}) as the motion model is made difficult by the non-linear  relationship between the position and the control variables. Conversely, in (\ref{nonhol2}), the position variables are simply affine functions of the control inputs. Consequently, a trajectory optimization based on (\ref{nonhol2}) is difficult due to the presence of non-linear curvature bounds. Thus, it can be seen that irrespective of which representation is used for the motion model of the non-holonomic vehicle, the trajectory optimization invariably takes the form of a difficult nonlinear and non-convex optimization problem. It is also interesting to note that if the curvature bounds are absent (e.g in case of differential drive robot), then using (\ref{nonhol2}), the problem of  connecting a pair of given positions/orientations in obstacle free space can be posed as a convex optimization problem. Further, if the obstacles (both static and dynamic) are approximated as circles, then the trajectory optimization with model (\ref{nonhol2}) can be shown to have a \emph{difference of convex} (DC) structure (e.g refer the derivations presented in \cite{sheuzi_shen_icra17}, \cite{aks_arxiv}, \cite{rafaello_scp}). Specially designed sequential convex programming (SCP) routines exist for solving optimization problems with DC structure \cite{boyd_sqp}. However, inclusion of curvature constraints in  (\ref{nonhol2}) destroys the \emph{differnece of convex} structure and consequently a more general non-linear optimization technique needs to be adopted \cite{alonzo_icra17}.

In this paper, we present a trajectory optimization which tries to  exploit as much as possible the problem structure. We advocate the use of representation (\ref{nonhol1}) as the model of the non-holonomic vehicle for the simple reason that herein, the curvature constraints are affine functions of control inputs. Thus, it is straightforward to ensure that the output of the trajectory optimization is always kinematically feasible. The central idea of the proposed work hinges on two inter related observations on nonlinear non-convex trajectory optimization. Firstly, the computational performance of such a trajectory optimization depends on how big the trust region is, which is the region where the  convex approximation of the problem holds. Secondly, on a related context, the reason why trajectory optimization with linear motion models and circular obstacles are shown to be easily solvable, atleast to local optimality is because they have infinitely big trust region owing to the DC structure of the problem \cite{boyd_sqp},\cite{sheuzi_shen_icra17}, \cite{aks_arxiv}, \cite{rafaello_scp}). The proposed trajectory optimization leverages both these observations by using better linearization for non-holonomic motion model and by reformulating collision avoidance between two convex polygons into that between a point and a circle.

In particular, the algorithmic contributions of the proposed work are two fold. Firstly, we present an alternating minimization routine which alternately operates in the space of angular accelerations and linear velocities. More precisely, the trajectory optimization has two separate layers wherein at the first layer, the angular accelerations are optimized while fixing the linear velocities. Subsequently, at the second layer, the linear velocities are optimized while the angular accelerations are fixed at the values obtained at the first layer and so on. Secondly, we use the concept of Minkowski sum and minimum bounding circle to reduce collision avoidance between two convex polygons to that between a point and a circle. These two contributions in in conjunction provides the proposed trajectory optimization with following key advantages over the current state of the art.

\begin{itemize}
\item Firstly, for a given angular acceleration profile, the non-holonomic motion model is  affine with respect to linear velocity. Further, fixing angular acceleration also ensures that the posture profile of the vehicle does not change in the velocity optimization layer. Since the Minkowski sum only depends on the relative posture of the vehicle and obstacle and not on their relative distance, the velocity optimization has the same DC structure which has been reported for linear systems with circular vehicles/obstacles \cite{sheuzi_shen_icra17}, \cite{aks_arxiv}, \cite{rafaello_scp}). We are not aware of any work which has highlighted such structure in non-holonomic trajectory optimization with convex polygon shapes.

\item We empirically show through extensive simulations that the proposed alternating optimization can afford  larger trust regions as compared to joint optimizations which simultaneously optimizes in the space of angular acceleration and linear velocity. Consequently, the alternating optimization  provides some gain in computation time over the joint formulations. 


\item The proposed circle approximation using the concept of Minkowski sum and minimum bounding circle leads to significantly less conservative approximation than representing each polygon individually through a circumscribing circle. Further, this is also an improvement over the approach presented in  \cite{alonzo_icra17} where polygons are represented as multiple overlapping circles. In particular, as compared to these cited works, the proposed approach  leads to reduced number of constraints in the trajectory optimization. For example, if three overlapping circles are used to represent just the vehicle polygon, then the trajectory optimization in \cite{alonzo_icra17} would have three times more number of constraints than the proposed approach.

\end{itemize}

\section{Problem Formulation}\label{traj_opt}
The trajectory optimization considered in this paper can be described by the following set of cost functions and constraints.

\begin{subequations}
\begin{align}
\arg\min_{\ddot{\theta}(t), v(t)} J = J_{smooth}+J_{terminal}.\label{cost}\\
X(t_{i+1}) = f(X(t_i), U(t_i)). \label{motion_mod}\\
v(t_i)\leq v_{max}. \label{velmax}\\
\dot{\theta}_{min} \leq \dot{\theta}(t_i)\leq \dot{\theta}_{max}. \label{thetadotmax}\\
a_{min}\leq \frac{v(t_{i+1})-v(t_i)}{\Delta t}\leq a_{max}.\label{accmax}\\
\ddot{\theta}_{min}\leq \ddot{\theta}(t_i)\leq \ddot{\theta}_{max}.\label{angularaccmax}\\
-\kappa_{max}v(t_i)\leq \dot{\theta}(t_i)\leq \kappa_{max}v(t_i).\label{curvmax}\\
C_{obst}(x(t_i), y(t_i), x_i(t_i), y_i(t_i), R_i)\leq 0.\label{collavoid}
\end{align}
\end{subequations}

\noindent Where, $X(t_i)= (x(t_i), y(t_i), \theta(t_i), \dot{\theta}(t_i))$ and represents the state of the system at time $t_i$. The individual cost terms can be represented in the following manner.

\begin{equation}
J_{smooth} = \sum_{i=1}^{N} \ddot{\theta}(t_i)^2+(\frac{v(t_{i-1})-2v(t_{i})+v(t_{i+1})}{\Delta t^2})^2
\label{cost_smooth}
\end{equation}

\begin{equation}
J_{terminal} = (x(t_N)-x_f)^2+(y(t_N)-y_f)^2
\label{cost_terminal}
\end{equation}

\noindent The objective  function (\ref{cost}) consists of  smoothness and terminal cost terms. As can be seen from (\ref{cost_smooth}), smoothness cost penalizes high value of angular accelerations and jerk modeled as second order finite difference of linear velocity. The terminal cost ensures that the optimal trajectory terminates as close as possible to the goal position $(x_f, y_f)$. The equality (\ref{motion_mod}) constrains the control variables and states to be compatible with the motion model of the robot. The inequality (\ref{velmax})-(\ref{thetadotmax}) represents bounds on linear and angular velocities while (\ref{accmax})-(\ref{angularaccmax}) can be thought as actuator constraints bounding the linear and angular acceleration magnitudes. The inequalities (\ref{curvmax}) are the curvature bound constraints. Inequalities (\ref{collavoid}) models the collision avoidance constraints and has the following algebraic form.

\begin{equation}
C_{obst}(.)\leq 0: -(x(t_i)-x_i(t_i))^2-(y(t_i)-y_i(t_i))^2+R_i^2\leq 0.
\label{coll_avoid_form}
\end{equation}

\noindent Where, $x_i(t_i) ,y_i(t_i), R_i$ are the position and size of the $i^{th}$ obstacle. For static obstacles, the position would be independent of $t_i$. The form of (\ref{coll_avoid_form}) assumes that the vehicle and the obstacles are both modeled as circular disks. As we show later, the same form can be leveraged to model collision avoidance between polygonal shapes as well.

The inequalities (\ref{coll_avoid_form}) are purely concave in terms of position variables $(x(t_i), y(t_i))$ or in other words has the so called \emph{difference of convex} form. Thus, as shown in \cite{boyd_sqp}, it can be upper bounded by the following affine inequality obtained by linearizing (\ref{coll_avoid_form}) by some initial guess trajectory $(^*x(t_i), ^*y(t_i))$. Satisfaction of (\ref{affine_approx}) ensures that the inequalities (\ref{coll_avoid_form}) are satisfied. 

\small
\begin{equation}
^{affine}C_{obst}(.)= {^*}C_{obst}+\bigtriangledown_x(x(t_i)-{^*}x(t_i))+\bigtriangledown_x(y(t_i)-{^*}y(t_i))
\label{affine_approx}
\end{equation}
\normalsize

\noindent The core complexity of optimization (\ref{cost})-(\ref{collavoid}) stems from the non-linear  motion model, (\ref{motion_mod}) since we have already constructed an affine approximation for collision avoidance constraints. If the motion model would have been affine, then the optimization could be efficiently  solved to (local) optimality through a specially designed sequential convex programming routine \cite{boyd_sqp}. In the case of non-linear motion model, the most common approach has been to adopt general non-linear optimization techniques wherein at each iteration, the non-linear motion model is approximated by an affine expression. In the next subsequent sections, we describe how the affine approximation, (\ref{affine_approx}) can be leveraged for modeling collision avoidance between polygonal shapes as well. We follow that by the presentation of our alternating minimization routine.

\section{Collision Avoidance Between Convex Polygons}
The proposed modeling approach for collision avoidance between convex polygons hinges on two basic ingredients namely Minkowski sum and circumscribing circle of an arbitrary polygon.

\subsection{Minkowski Sum}
\noindent Minkowski sum of two sets $P, Q \in \Re^n$ can be defined as

\begin{equation}
P\oplus Q = \{p+q\vert p\in P, q \in Q\}
\label{minkowski}
\end{equation} 

\noindent In our case, $P,Q$ are polygons in $\Re^2$. Minkowski sum boundary is similar to contact space, which means robot is placed in contact with obstacles but with out collision \cite{hybrid_minkowski_planner}. The boundary of the Minkowski polygon can be represented as

\begin{equation}
L = Q+(-P)
\label{minkowski_boundary}
\end{equation}

\noindent Thus, using $L$ one can replace one of the polygons to a point and the other to an arbitrary polygon. The Minkowski sum of two convex polygons with $n$ sides has a computational complexity of $O(n^2)$ and thus can be computed with relative ease for shapes like rectangles, squares.

\begin{figure}[!tbh]
  \centering
   \subfigure[]{
    \includegraphics[width= 2in, height= 2in] {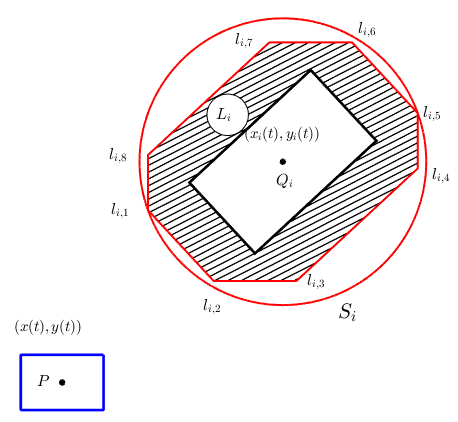}
    \label{min_circle}
   }
   \caption{Shaded region in the figure shows minkowski sum between $P$ and $Q_i$ and a bounding circle $S_i$(in red) around this sum}          
\end{figure}

\subsection{Minimum Bounding Circle}
\noindent Given a set of vertex points, $L_i$ of the Minkowski polygon $L$, we compute the smallest circle containing the polygon through the concept of minimum bounding circle. Various algorithms e.g based on randomization and quadratic programming are generally used to find such circle fitting into vertices of $L_i$.  The computational complexity is linear with respect to number of vertices, $L_i$. We use the open source implementation \cite{matlab_minbound} in our work.

\section{Alternating Optimization}

Algorithm \ref{algo1} summarizes the proposed alternating optimization routine. It starts with (line 1) choosing an initial guess for linear velocity, ${^*}v(t)$, angular acceleration, $^*\ddot{\theta}$ and a counter $k$ along with two positive weights, $w_{\theta}, w_{v}$. The function $InitialTraj(.)$ (line 3) computes an initial guess for position and heading trajectory, $(^*x^k(t), ^*y^k(t), ^{*}\theta^k(t))$ based on the guess for linear velocity and acceleration. Lines 6 and 12 represent the angular acceleration and linear velocity optimization layer respectively. Both the layers continue till the change in the cost function between subsequent iteration is greater than the threshold, $\epsilon$ and the collision avoidance constraints are not satisfied. The optimal solution obtained after each layer is used to update the initial guesses of angular accelerations and linear velocities (lines 11 and 17).

\subsection{Angular Acceleration Layer}
\noindent The angular acceleration layer is obtained by extracting the $\theta(t)$ dependent terms from the optimization (\ref{cost})-(\ref{collavoid}). The following points are worth pointing out here

\begin{itemize}
\item Firstly, note the motion model $f^{\theta}(.)$ which is obtained by first order Taylor series expansion of the first two equations in (\ref{nonhol1}) around $^{*}\theta(t)$. Consequently, we obtain a motion model which is affine with respect to $\theta$  and consequently angular acceleration, $\ddot{\theta}(t)$.

\item  The affine approximation  holds only  in the vicinity of $^{*}\ddot{\theta}^k(t)$. Thus, a trust region needs to be incorporated to ensure that $^{*}\ddot{\theta}^k(t)$ and $^{*}\ddot{\theta}^{k+1}(t)$ are sufficiently close to each other. The last inequality in line 6 of algorithm \ref{algo1} which puts a box constraints on $\ddot{\theta}(t)$  serve this purpose. The trust region is modified at each iteration of  based on constraint violations as discussed in \cite{boyd_sqp}.

\item The collision avoidance constraints have been augmented with a non-negative slack variable $s_{\theta}(t)$. This is to ensure that the algorithm \ref{algo1} continues to make progress towards the optimal solution even if the initial guess trajectory $(^*x^k(t), ^*y^k(t))$ renders $^{affine}C(.)$ infeasible. Consequently, we also incorporate a penalty on the slack variables in the cost function. The weights of the penalty is sequentially increased as long as $C_{obst}(.)>0$. 
\end{itemize}

\subsection{Linear Velocity Layer}
\noindent This layer has only such terms from the cost and constraint functions from (\ref{cost})-(\ref{collavoid}) which explicitly depends on the linear velocity. The following key points should be noted.

\begin{itemize}
\item Note, the motion model, $f^{v}(.)$ which has been obtained from (\ref{nonhol1}) by fixing the $\theta(t)$ based on the angular acceleration, $\ddot{\theta}(t)^{k+1}$ obtained at the previous layer. Consequently, $f^{v}(.)$ is affine with respect to linear velocity.

\item The naturally affine motion model means that there is no need to incorporate trust region constraints in this layer.

\item The collision avoidance constraints are augmented with non-negative slacks $s_{v}(t)$ similar to angular acceleration layer. The penalty on the slacks also follows the same reasoning.
\end{itemize}

\subsection{A Note on Structure}
\noindent The linear velocity optimization layer in algorithm  (\ref{algo1}) has the same DC structure as that reported in \cite{sheuzi_shen_icra17}, \cite{aks_arxiv}, \cite{rafaello_scp}). Consequently, the computational performance of algorithm (\ref{algo1}) depends on how big the trust region at the angular acceleration optimization layer is.

\begin{algorithm}
 \caption{Alternating Optimization}\label{algo1}
    \begin{algorithmic}[1]
    \State  \textbf{Initialization}: Initial guess for ${^*}v(t)$, $^*\ddot{\theta}(t)$, iteration counter, $k=0$, $w_{\theta}, w_{v}$ \\.
\State $(^*x^k(t), ^*y^k(t))=InitialTraj({^*}v(t)$, $^*\ddot{\theta}(t))$.      \\
     \While {$\vert J^{k+1}_{\theta}-J^{k}_{\theta}\vert \geq \epsilon$ and $\vert J^{k+1}_{v}-J^{k}_{v}\vert \geq \epsilon$} and  $C_{obst}\geq 0$ 
    \State 
\begin{subequations}
\begin{align}
\ddot{\theta}(t)^{k+1} = \arg\min J_{\theta} + \sum_{i=1}^{N} \ddot{\theta}(t_i)^2+\sum w_{\theta}s_{\theta}.\nonumber \\
X(t_{i+1}) = f^{\theta}(X(t_i), U(t_i)). \nonumber \\
\dot{\theta}_{min} \leq \dot{\theta}(t_i)\leq \dot{\theta}_{max}. \nonumber \\
\ddot{\theta}_{min}\leq \ddot{\theta}(t_i)\leq \ddot{\theta}_{max}\nonumber\\
-\kappa_{max}{^*}v(t_i)\leq \dot{\theta}(t_i)\leq \kappa_{max}v{^*}(t_i).\nonumber \\
^{affine}C_{obst}({^{k}}x(t_i), {^k}y(t_i))-s_{\theta}(t_i)\leq 0.\nonumber\\
s_{\theta}(t_i)\geq 0. \nonumber\\
-\ddot{\theta}^{trust}(t_i)\leq \ddot{\theta}(t_i) \leq \ddot{\theta}^{trust}(t_i).\nonumber  
\end{align}
\end{subequations}
\If{$C_{obst}(.)>0$}
\State $w_{\theta} \leftarrow w_{\theta}*\delta$
\EndIf \\   
\State $^{*}\ddot{\theta}(t)\leftarrow \ddot{\theta}(t)^{k+1}$     
    \State 
\begin{subequations}
\begin{align}
v(t)^{k+1} = \arg\min J_{v} +\sum w_{v}s_{v}.\nonumber \\
X(t_{i+1}) = f^{v}(X(t_i), U(t_i), \ddot{\theta}(t)^{k+1}). \nonumber \\
v(t_i)\leq v_{max}. \nonumber\\
a_{min}\leq \frac{v(t_{i+1})-v(t_i)}{\Delta t}\leq a_{max}.\nonumber\\
-\kappa_{max}v(t_i)\leq {^*}\dot{\theta}^k(t_i)\leq \kappa_{max}v(t_i).\nonumber \\
^{affine}C_{obst}(.)-s_{v}(t_i)\leq 0.\nonumber\\
s_{v}(t_i)\geq 0. \nonumber
\end{align}
\end{subequations}
\If{$C_{obst}({^{k}}x(t_i), {^k}y(t_i), ..)>0$}
\State $w_{v} \leftarrow w_{v}*\delta$
\EndIf \\   
\State$^{*}v(t)\leftarrow v(t)^{k+1}$
\State$k \leftarrow k+1$
    \EndWhile  
    
        \end{algorithmic}  
        \normalsize 
        \end{algorithm}

\section{IMPLEMENTATION AND RESULTS}

To compare alternating minimization with an more general approach where linear velocity and angular acceleration are optimized at the same time (Joint Minimization), we have prototyped both of them in CVX \cite{cvx_matlab}. Then, comparisons were made on runtime, number of iterations taken to converge, arc-length of path generated, velocity profile smoothness, acceleration profile smoothness, angular acceleration profile smoothness. In subsequent subsection we discuss details of these comparisons and results obtained by implementing our approach in MPC framework. By testing our approach on some typical urban overtaking and merging scenarios with dynamic obstacles we show robustness of our approach.

\subsection{Comparisons}

\begin{figure}[H]
  \centering
   \subfigure[]{
    \includegraphics[width= 1.5in, height=3.0cm] {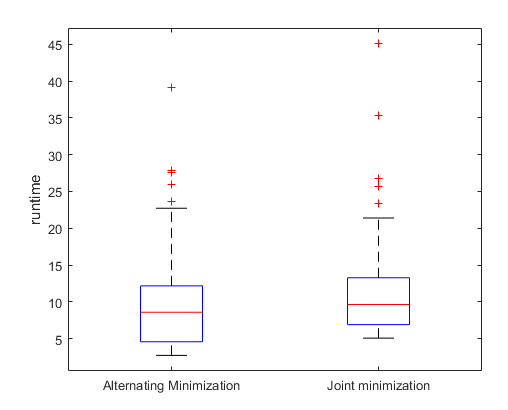}
    \label{runtime}
   }
   \subfigure[]{
    \includegraphics[width= 1.5in, height=3.0cm] {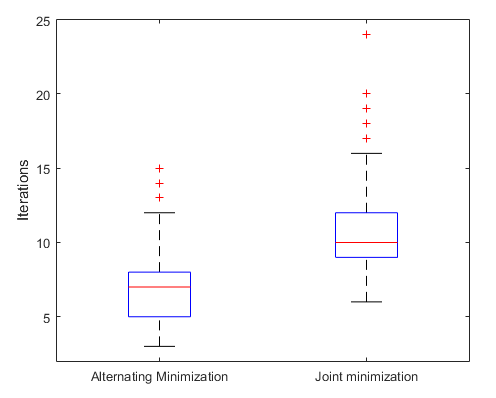}
    \label{iterations}
   }
   \caption{Comparisons showing runtime and iterations of joint minimization and alternating minimization. we observe that our approach has improved runtime by $1.8sec$ and iterations reduced to almost half }          
\end{figure}

The comparison between runtime and iterations of alternating and joint minimization in Fig. \ref{runtime}-\ref{iterations} show that our approach has half the number of iterations to that of joint minimization. We also notice an improvement in runtime by $1.8s$. Absence of trust region in velocity layer has provided some advantage. We can expect a further improvement in runtime by only iterating over one layer once change in other layer is less than threshold.

\begin{figure}[H]
  \centering
   \subfigure[]{
    \includegraphics[width= 1.5in, height=3.0cm] {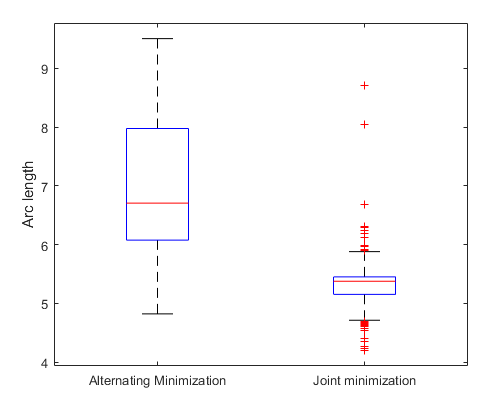}
    \label{arclength}
   }
   \subfigure[]{
    \includegraphics[width= 1.5in, height=3.0cm] {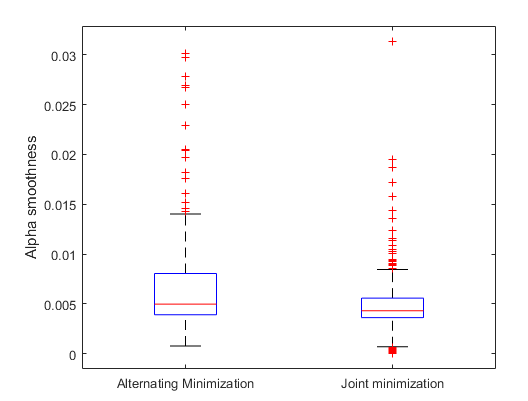}
    \label{alpha}
   }
      \subfigure[]{
    \includegraphics[width= 1.5in, height=3.0cm] {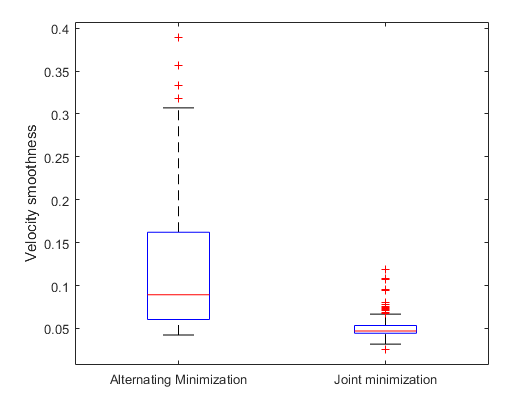}
    \label{velocity}
   }
   \subfigure[]{
    \includegraphics[width= 1.5in, height=3.0cm] {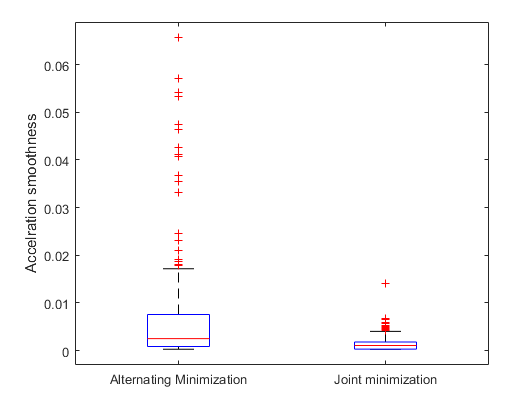}
    \label{acceleration}
   }
   \caption{Comparisons showing smoothness of different control variables compared between alternating minimization and joint minimization}          
\end{figure}

Fig. \ref{arclength}-\ref{acceleration} show comparison between different smoothness terms. Smoothness of a control variable is defined as sum of square of rate of change of that variable. In our comparison joint minimization has better smoothness due to its search in angular acceleration space and linear velocity space at the same time. However, the changes in control variables are strictly under the limits provided in inequalities in line 6 and 12 of Algorithm \ref{algo1} and hence are allowed with in the dynamic limits of model. 

\subsection{Model Predictive Control}

To evaluate our approach, we have used cvxgen \cite{cvxgen_olf} which generates appropriate C code to be mexed with MATLAB to speed up the process of optimization. During implementation into MPC framework we have taken a planning horizon of 5 seconds, for 50 steps with a $\delta T$ of 0.1 seconds. We replan at a frequency of 5hz and also limited to a maximum speed of vehicle to 10m/s. Below is box plot of runtime of cvxgen, Though worst case comes out to be around 800ms which is a scenario with parking lot. This is being further improved by including only active sets of constraints.
\begin{center}
\includegraphics[width=2.1in]{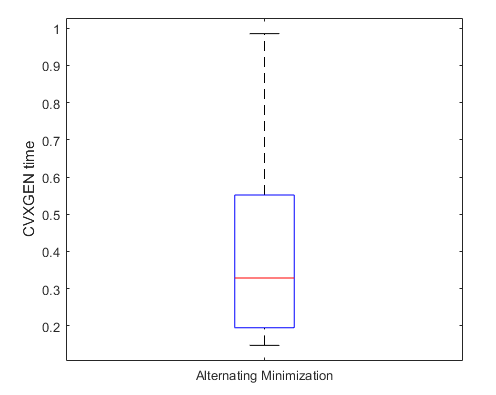}
\end{center}

\begin{figure}[H]
  \centering
   \subfigure[]{
    \includegraphics[width= 3in, height=3.0cm] {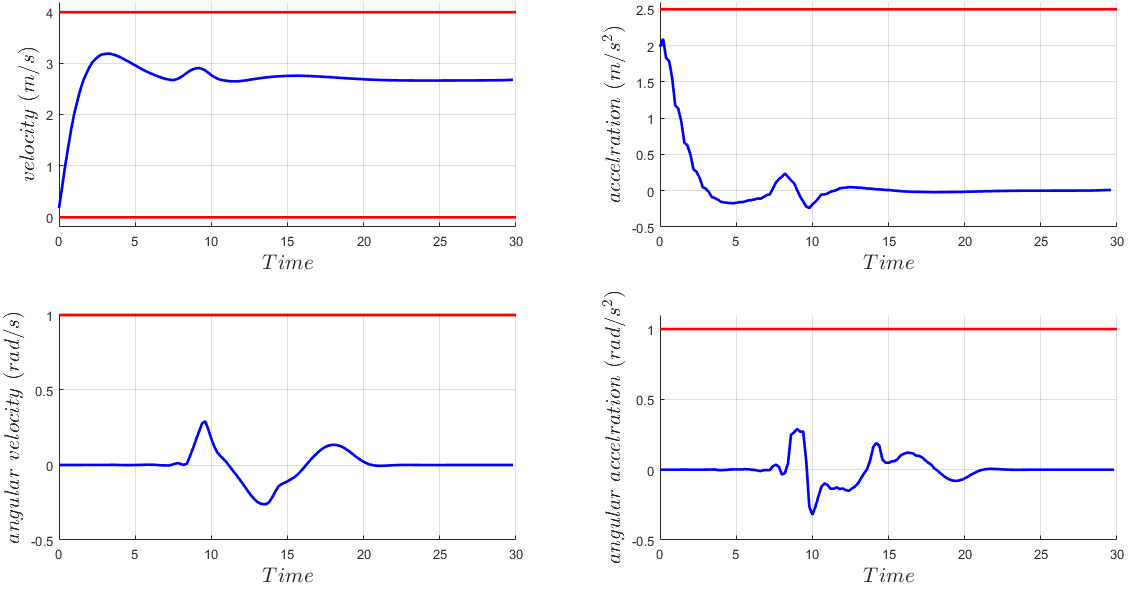}
    \label{overtaking_1}
   }
   \caption{This plot shows linear velocity, acceleration, angular velocity and angular acceleration in four subplots}          
\end{figure}

	In addition, we have implemented few maneuvers in GAZEBO \cite{gazebo} and ROS \cite{ros}. The control points input for the MPC framework comes from cruise control with an average velocity of 2.7m/s equally spaced along the lane. The red lines in Fig.\ref{overtaking_1} indicate maximum and minimum bounds provided to control variables during optimization.

\begin{figure}[H]
  \centering
   \subfigure[]{
    \includegraphics[width= 3in, height=0.8cm] {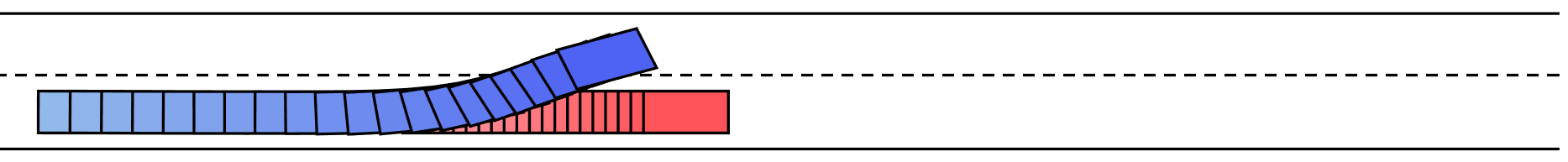}
    \label{overtaking_2}
   }\\
   \subfigure[]{
    \includegraphics[width= 1.5in, height=3.0cm] {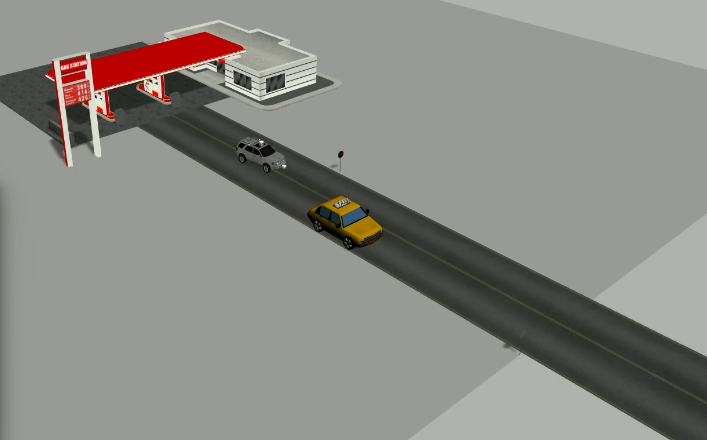}
    \label{overtaking_3}
   }\\
      \subfigure[]{
    \includegraphics[width= 3in, height=0.8cm] {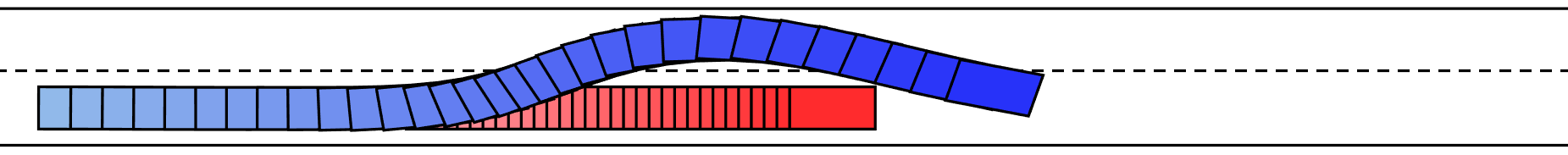}
    \label{overtaking_4}
   }\\
   \subfigure[]{
    \includegraphics[width= 1.5in, height=3.0cm] {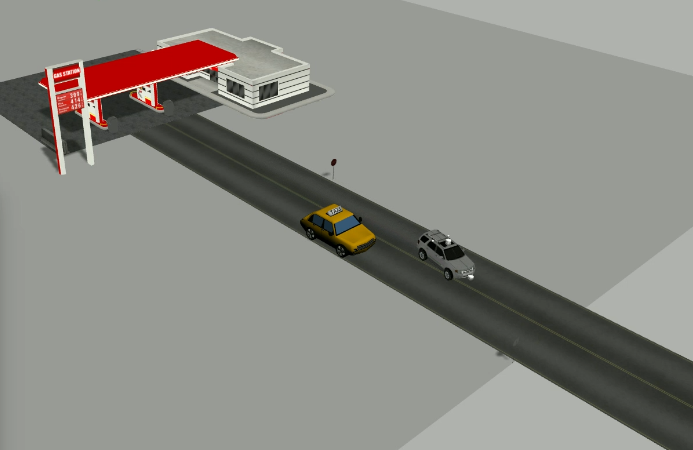}
    \label{overtaking_5}
   }
   \caption{Overtaking manuver simulated in Gazebo}          
\end{figure}

Initially, vehicle starts from rest and accelerates up to 2.7m/s to track the way points on the path. Later during overtaking maneuver, we observed an increase in velocity and gain in angular velocity (between 7s to 12s). We also notice control variables with in their dynamic limits provided at the same time completing overtaking maneuver. In Fig \ref{overtaking_2}-\ref{overtaking_5}, we can observe a sequence of overtaking maneuver. Figures \ref{overtaking_2} and \ref{overtaking_4} shows vehicle location and obstacle location every 100ms  with colour change from light blue to dark blue.

In a more general scenario, where our ego vehicle is trying to overtake slow moving vehicle on our lane. However, we find another slow moving vehicle obstructing overtaking maneuver on other lane. In such case, we observe our vehicle slowing down on current lane and then speeding up as soon as vehicle on other lane vehicle passes by.

\begin{figure}[H]
  \centering
   \subfigure[]{
    \includegraphics[width= 2.6in, height=1.2cm] {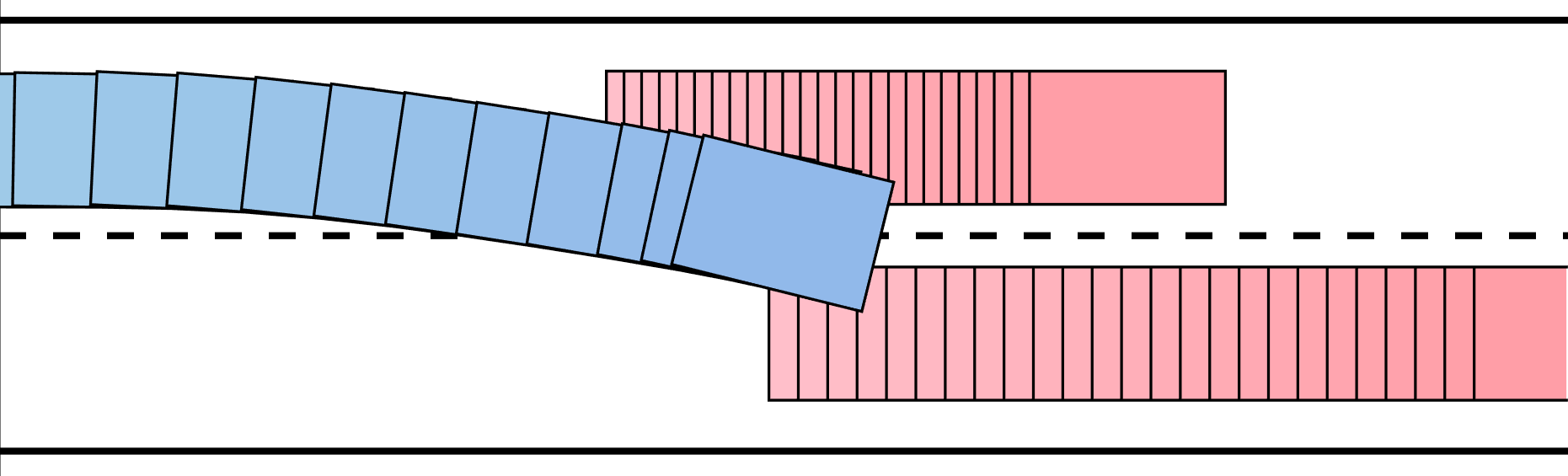}
    \label{compli_1}
   }\\
   \subfigure[]{
    \includegraphics[width= 1.5in, height=3.0cm] {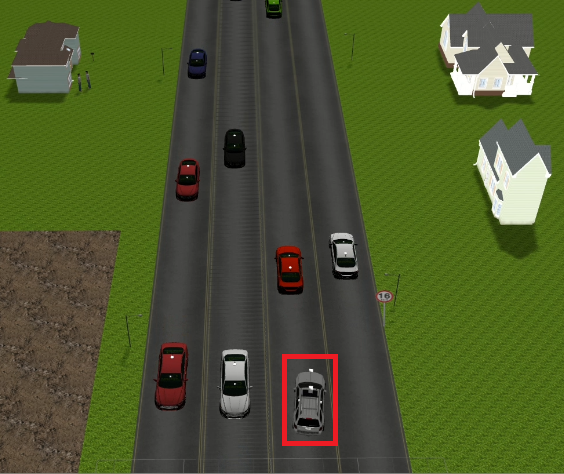}
    \label{compli_2}
   }\\
      \subfigure[]{
    \includegraphics[width= 2.6in, height=1.2cm] {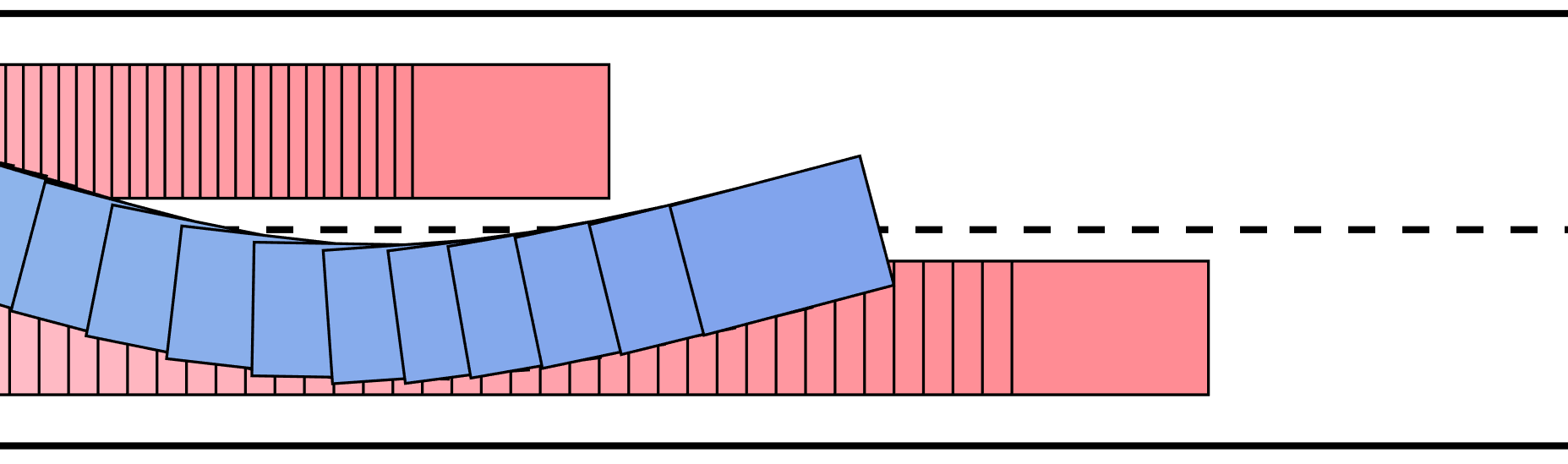}
    \label{compli_3}
   }\\
   \subfigure[]{
    \includegraphics[width= 1.5in, height=3.0cm] {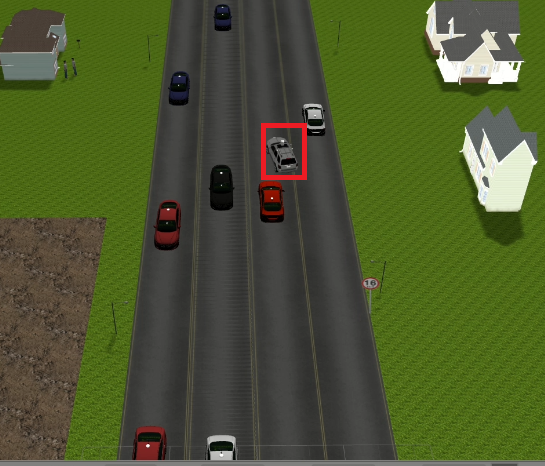}
    \label{compli_4}
   }\\
      \subfigure[]{
    \includegraphics[width= 2.6in, height=1.2cm] {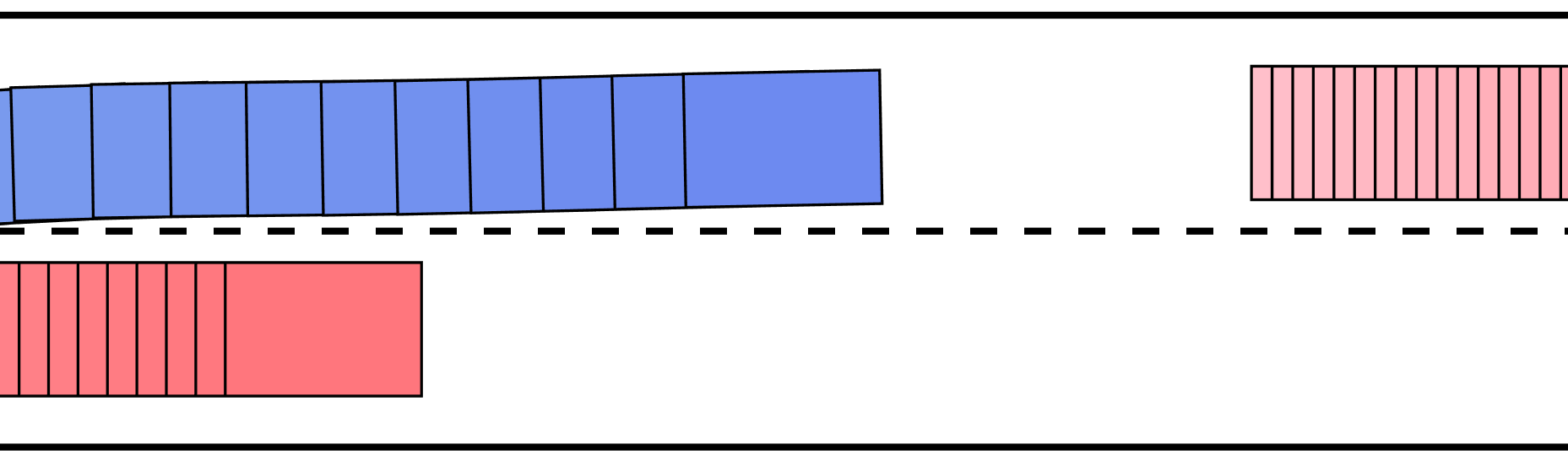}
    \label{compli_5}
   }\\
   \subfigure[]{
    \includegraphics[width= 1.5in, height=3.0cm] {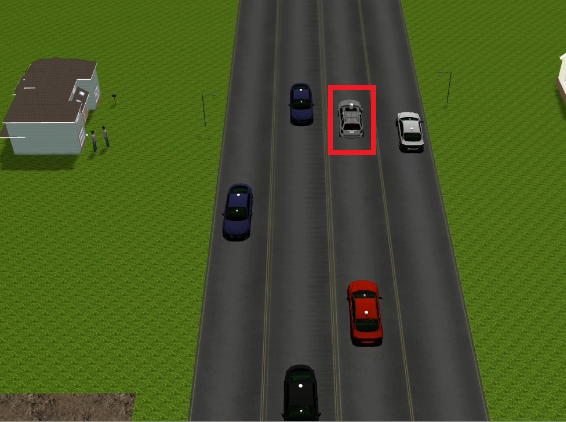}
    \label{compli_6}
   }
   \caption{Overtaking manuver with slow moving vehicle on other lane simulated in Gazebo}          
\end{figure}

Here vehicle encircled in red is ego vehicle and we are trying to overtake slow moving red car on our lane and other lane has white car that obstructs our maneuver.Figures \ref{compli_1},\ref{compli_3} and \ref{overtaking_5} shows vehicle location and obstacle location every 100ms  with colour change from light blue to dark blue.

\begin{figure}[H]
  \centering
   \subfigure[]{
    \includegraphics[width= 1.5in, height=3cm] {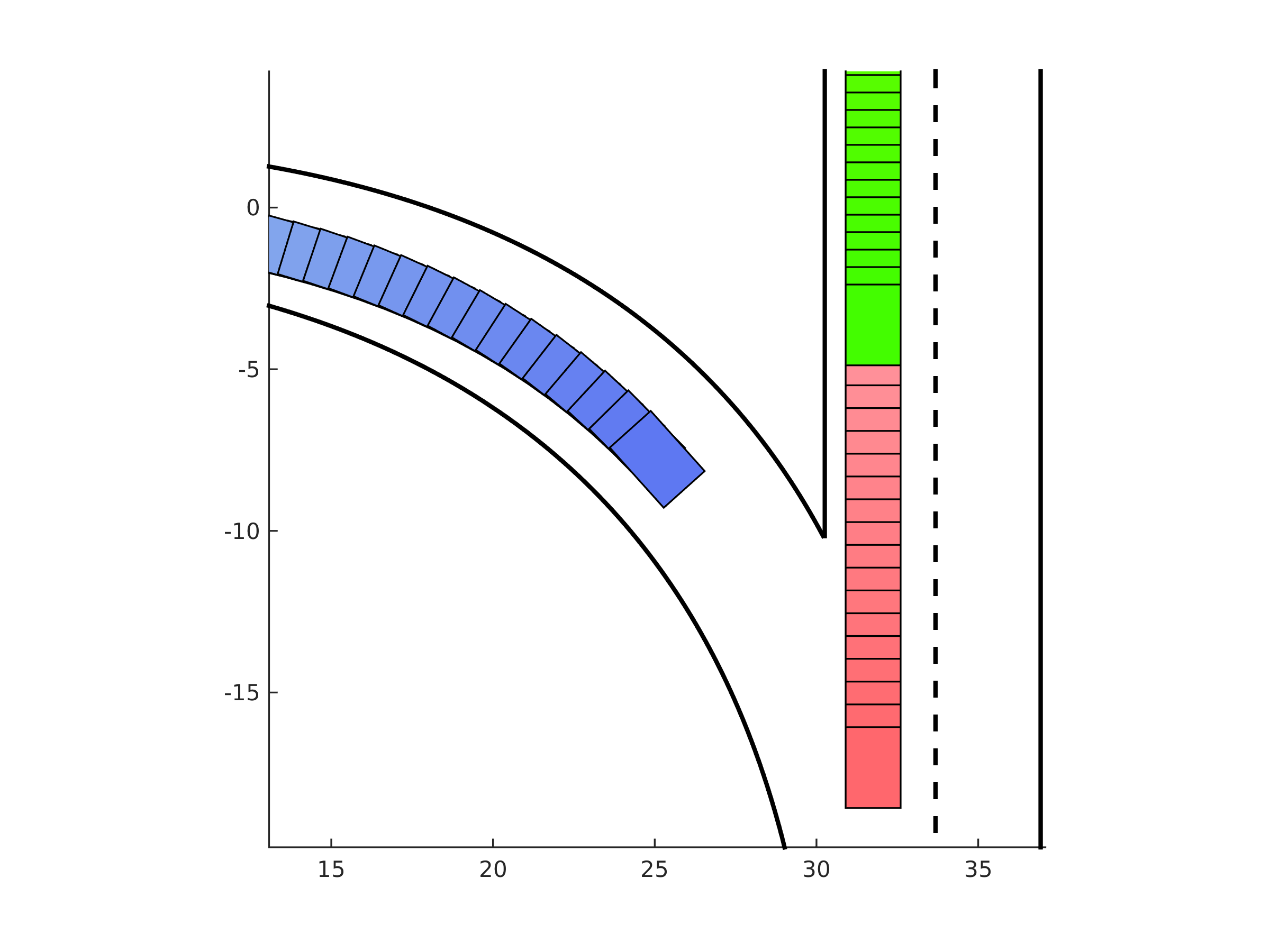}
    \label{compli_11}
   }
   \subfigure[]{
    \includegraphics[width= 1.5in, height=3.0cm] {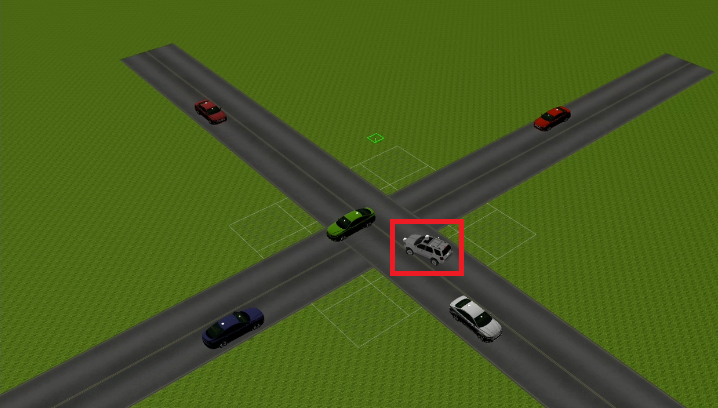}
    \label{compli_12}
   }
      \subfigure[]{
    \includegraphics[width= 1.5in, height=3cm] {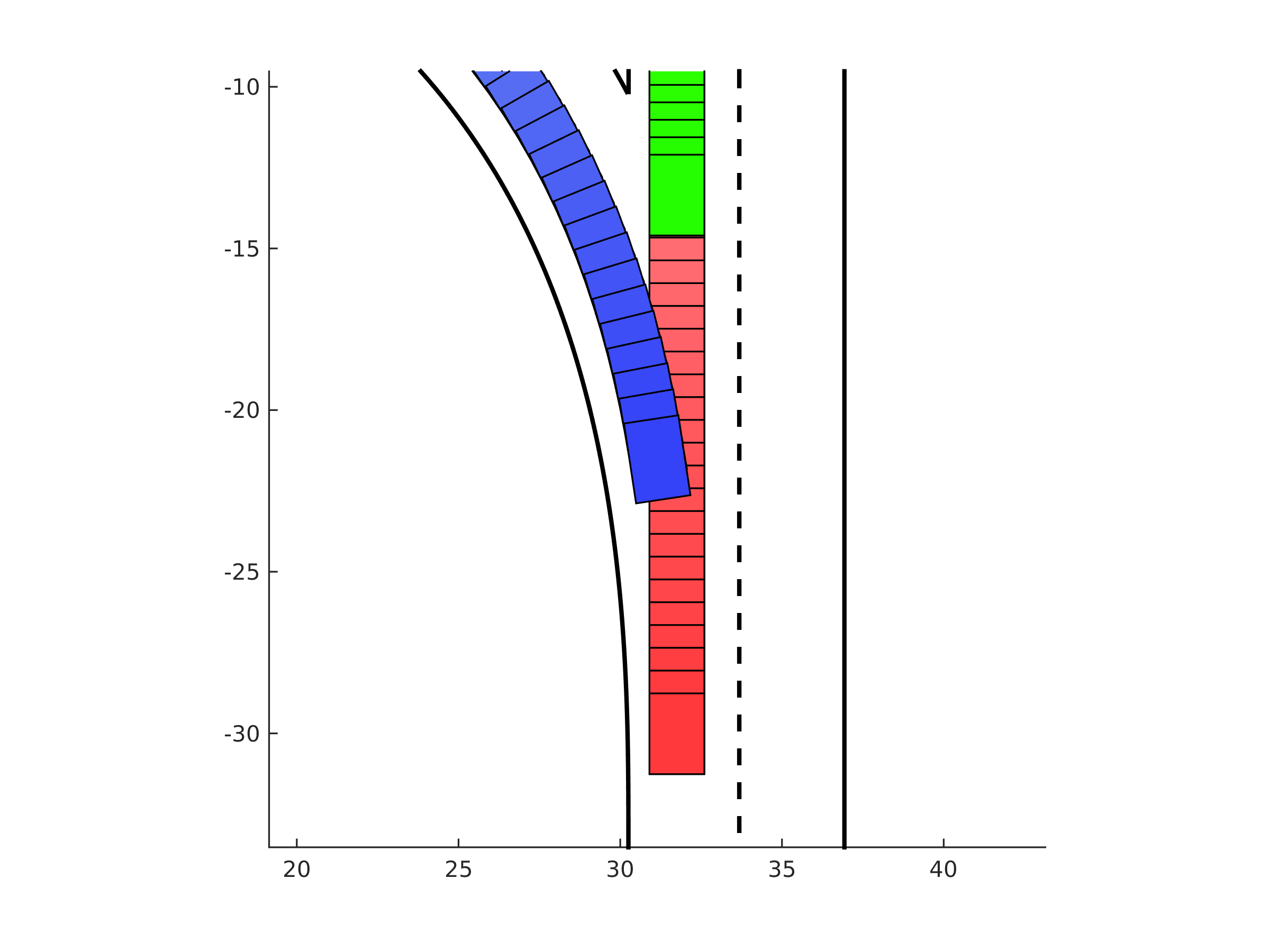}
    \label{compli_13}
   }
   \subfigure[]{
    \includegraphics[width= 1.5in, height=3.0cm] {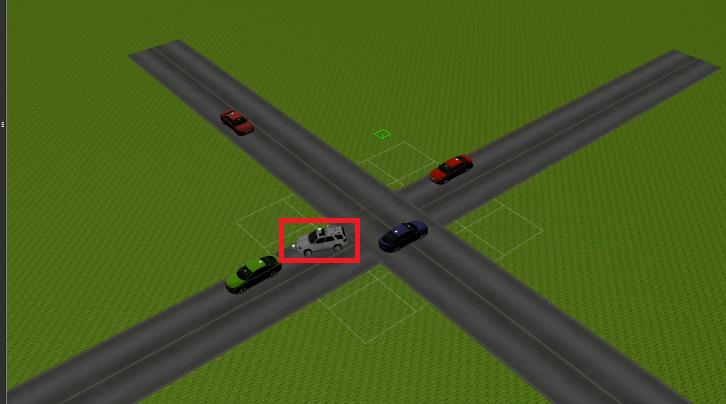}
    \label{compli_14}
   }
      \subfigure[]{
    \includegraphics[width= 1.5in, height=3cm] {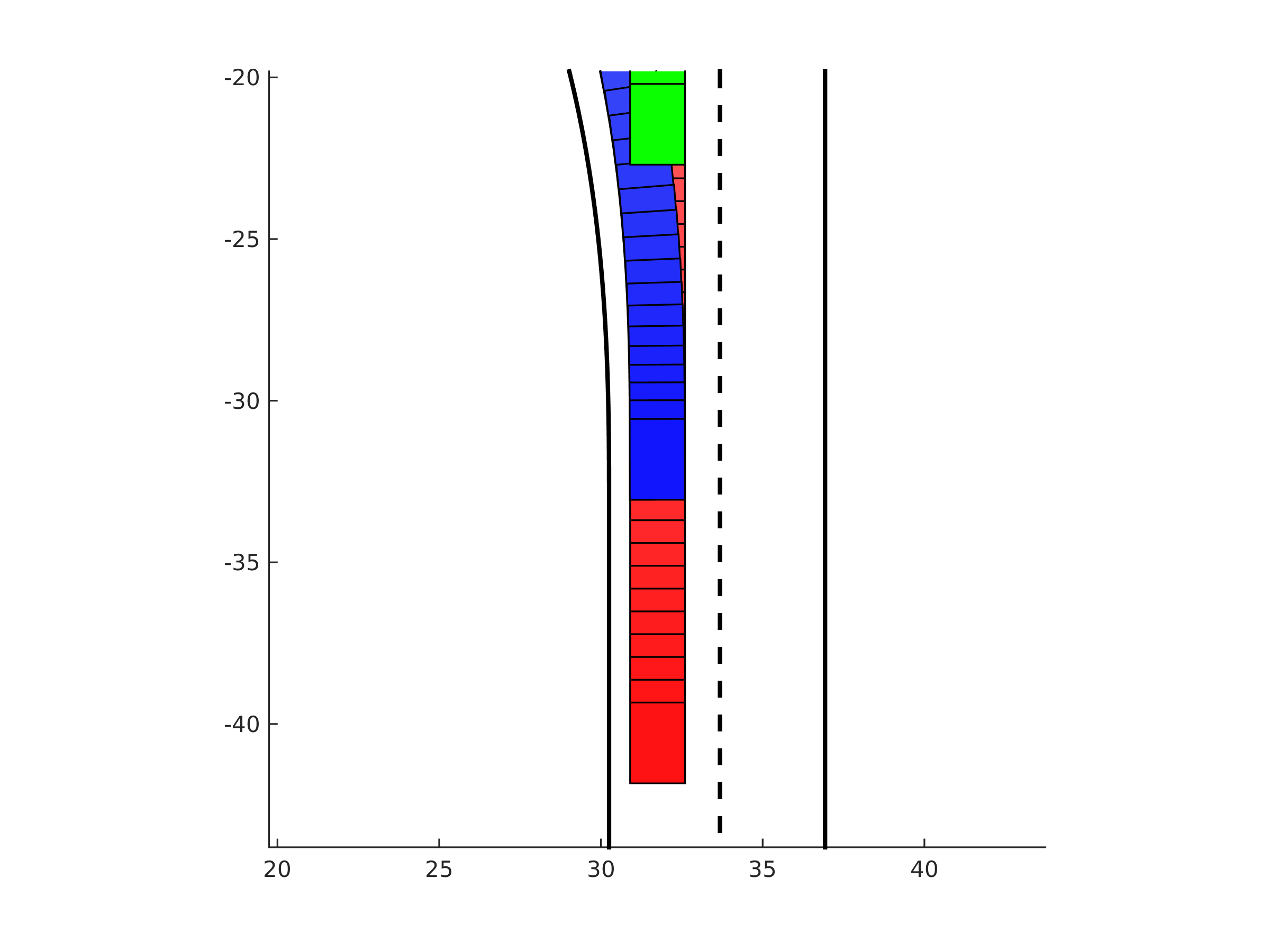}
    \label{compli_15}
   }
   \subfigure[]{
    \includegraphics[width= 1.5in, height=3.0cm] {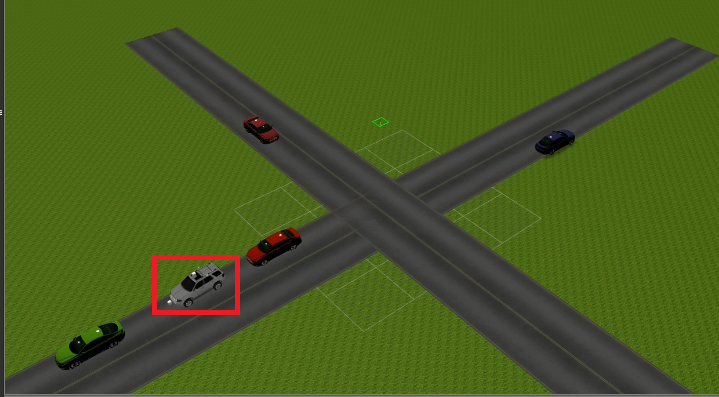}
    \label{compli_16}
   }
   \caption{Merging maneuver simulated in Gazebo}          
\end{figure}

Similarly we have simulated merging maneuver in gazebo, in which our vehicle(marked with red box) merges into other vehicle at an intersection. During this scenario, the car moving ahead obstructs our simple cruise control path. Then it is observed that ego vehicle reduces its speed and merged in between red and green cars marked in \ref{compli_11}, \ref{compli_13} and \ref{compli_15}.

\begin{figure}[H]
  \centering
   \subfigure[]{
    \includegraphics[width= 1.5in] {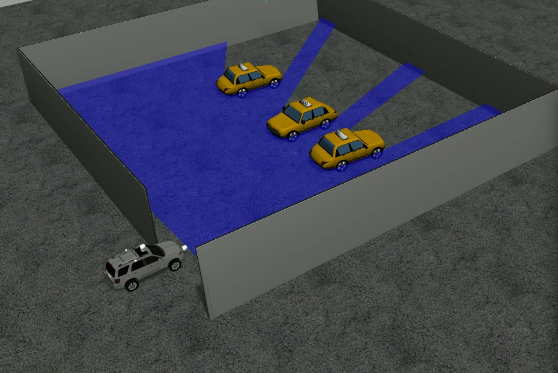}
    \label{compli_111}
   }
   \subfigure[]{
    \includegraphics[width= 1.5in] {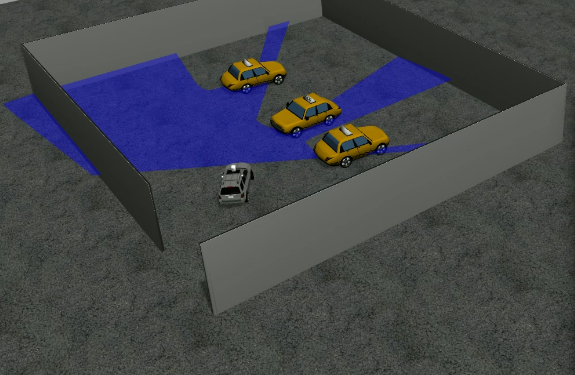}
    \label{compli_112}
   }
      \subfigure[]{
    \includegraphics[width= 1.5in] {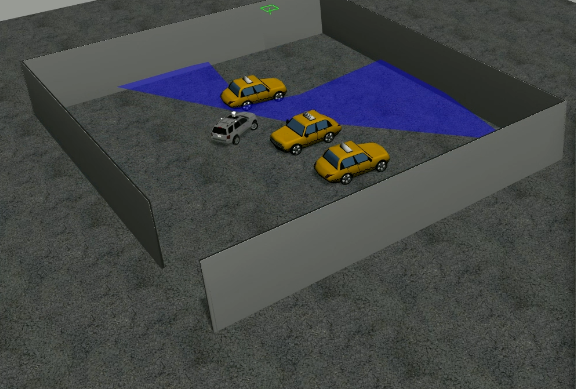}
    \label{compli_113}
   }
   \subfigure[]{
    \includegraphics[width= 1.5in] {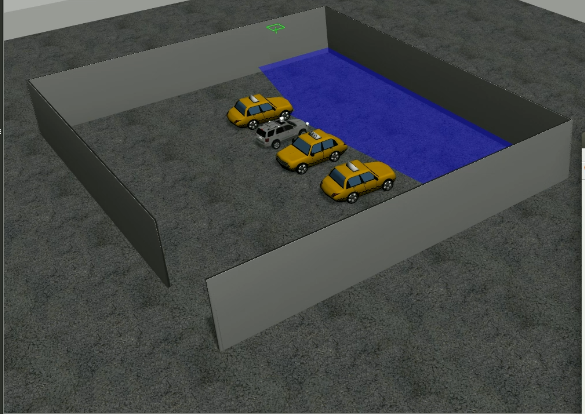}
    \label{compli_114}
   }
   \caption{Maneuver of ego vehicle in a parking lot simulated in Gazebo}          
\end{figure}

In this paper, we have also tested our MPC framework in an unstructured environments such as parking lot. Figures \ref{compli_111} - \ref{compli_114}
show one such scenario in a parking lot with space between yellow taxis parked in.

\bibliographystyle{IEEEtran}  
\bibliography{ref}

\end{document}